\documentclass{article}

\usepackage{PRIMEarxiv}

\usepackage[utf8]{inputenc} 
\usepackage[T1]{fontenc}    
\usepackage{hyperref}       
\usepackage{url}            
\usepackage{booktabs}       
\usepackage{amsfonts}       
\usepackage{nicefrac}       
\usepackage{microtype}      
\usepackage{lipsum}
\usepackage{fancyhdr}       
\usepackage{graphicx}       
\graphicspath{{media/}}     
\usepackage{natbib}
\usepackage{graphicx}
\usepackage[utf8]{inputenc} 
\usepackage[T1]{fontenc}    
\usepackage{hyperref}       
\usepackage{url}            
\usepackage{booktabs}       
\usepackage{amsfonts}       
\usepackage{nicefrac}       
\usepackage{microtype}      
\usepackage{xcolor}         
\usepackage{amsmath}
\usepackage{todonotes}
\title{Improved deep learning of chaotic dynamical systems with multistep penalty losses
}
\author{%
  Dibyajyoti Chakraborty\thanks{Corresponding author: d.chakraborty@psu.edu} \\
  School of Information Sciences and Technology\\
  Pennsylvania State University\\
  University Park, PA-16802, USA.  \\
  \And
  Seung Whan Chung \\
   Center for Applied Scientific Computing \\
   Lawrence Livermore National Laboratory \\
  Livermore, CA-94550, USA \\
  \AND
  Ashesh Chattopadhyay \\
  Department of Applied Mathematics \\
  University of California Santa Cruz \\
  1156 High Street, Santa Cruz, CA-95064, USA\\
  \And
  Romit Maulik \\
  School of Information Sciences and Technology \\
  Pennsylvania State University \\
  University Park, PA-16802, USA.  \\
}

\begin{document}
\maketitle

\begin{abstract}
Predicting the long-term behavior of chaotic systems remains a formidable challenge due to their extreme sensitivity to initial conditions and the inherent limitations of traditional data-driven modeling approaches. This paper introduces a novel framework that addresses these challenges by leveraging the recently proposed multi-step penalty (MP) optimization technique. Our approach extends the applicability of MP optimization to a wide range of deep learning architectures, including Fourier Neural Operators and UNETs. By introducing penalized local discontinuities in the forecast trajectory, we effectively handle the non-convexity of loss landscapes commonly encountered in training neural networks for chaotic systems. We demonstrate the effectiveness of our method through its application to two challenging use-cases: the prediction of flow velocity evolution in two-dimensional turbulence and ocean dynamics using reanalysis data. Our results highlight the potential of this approach for accurate and stable long-term prediction of chaotic dynamics, paving the way for new advancements in data-driven modeling of complex natural phenomena.
\end{abstract}


\section{Introduction}
\vspace{-0.2cm}
Chaotic systems are ubiquitous in nature, encompassing fields as diverse as meteorology, fluid dynamics, and chemical reactions. They exhibit complex multi-scale dynamics, without any scale separation, making their forecasts extremely challenging. They are also characterized by their extreme sensitivity to initial conditions, meaning a small perturbation in the initial condition leads to completely diverging trajectories over time. A deterministic long-term prediction for chaotic systems is irrelevant due to their very nature. Therefore, several current works on data driven long term prediction of chaotic systems focus on preserving the invariant statistics of the system\citep{linot2023stabilized,schiff2024dyslim,li2021markov,guan2024lucie}. However, minimizing the deviations from ground truth in one autoregressive time step of prediction using a mean-squared error, typically used to optimize ML models, is not effective for long term dynamics. Recent developments have tried to tackle this limitation by several techniques like using multiple timesteps for accumulating errors before gradient computation \citep{keisler2022forecasting}, including structures from governing differential equations\citep{linot2023stabilized} and implementing physical laws in optimization \citep{raissi2019physics}.
\par

In this work, we focus on the challenges data-driven ML models trained using multiple timesteps (rollouts) face for predicting chaotic systems.
Gradient-based optimization used in neural networks, aiming to minimize the difference between predictions and ground truth, proves particularly difficult for such systems. The extreme sensitivity to perturbations leads to exploding gradients during optimization for any long term objective with underlying chaotic dynamics \citep{lea2000sensitivity}. Additionally, even a theoretically convex objective function becomes highly non-convex in numerical implementation when involving long chaotic trajectories, often trapping the optimization process in sub-optimal local minima \citep{chung2022optimization}. For more details on non-convexity and loss landscape we refer readers to \cite{chakraborty2024divide}. This challenge shares similarities with the well-known exploding/vanishing gradient problem in deep learning \citep{hanin2018neural,philipp2017exploding}. Similar to how chaotic dynamics evolve, the repeated application of deep neuron layers can cause extreme gradients while automatic differentiation, obstructing the training process. Although there are some previous works by the machine learning community(Refer section 2 in \cite{chakraborty2024divide}) trying to tackle this issue, it is still an open problem and an area of active research. 
\par
A solution to the exploding gradients problem, proposed by \cite{blonigan2014least} is the Lease Square Shadowing (LSS) method. They use the shadowing lemma, which states that there exists a trajectory that always stays close to the reference trajectory with slightly perturbed initial conditions. The brute-force computation of shadowing lemma requires a cubic cost with respect to the number of parameters rendering it unusable for deep learning. \cite{chung2022optimization} introduced the multi-step penalty(MP) optimization which uses segmented time intervals and introduces penalized local discontinuities to optimize the objective along with minimizing the discontinuities. As an alternative to LSS method, \cite{chakraborty2024divide} showed that the MP optimization can reduce the gradient computation cost from cubic to linear with respect to the number of parameters. They implemented it on several chaotic systems like Lorenz, 2D turbulence and weather to achieve long term stability. However, the dynamical core of their work was based specifically on the Neural Ordinary Differential Equations \citep{chen2018neural}. In this paper, we propose a modified extension of the MP optimization algorithm to other deep learning algorithms that can predict dynamics autoregressively, thereby assessing the general applicability of the optimization technique. We implement it on two popular deep learning architectures for dynamical systems, namely the Fourier Neural Operator(\cite{li2020fourier}) and UNET (\cite{ronneberger2015u}). We focus on two chaotic dynamical systems - High Reynolds number (Re $\sim10^5$) 2D turbulence with Kolmogorov forcing and the northwest Atlantic Ocean western boundary current (the Gulf Stream) dataset introduced in \cite{chattopadhyay2024oceannet}. 
\vspace{-0.4cm}
\section{Methodology}
\vspace{-0.2cm}
Let us consider an operator $S$ that advances the state $q$ of a dynamical system in time. It can be considered as the theoretical solution of any underlying governing differential equation that controls the dynamics of a system. So, the state evolution can be given as,
\begin{equation}
    q_t = S(q_{t-1}) = S(S(S(...S(q_0)))) = S^{t}(q_0)
\end{equation}
where $q_t$ is the state at time $t$. $S$ can be approximated by a neural network architecture $F_\theta(q)$, depending on learnable parameters $\theta$. The parameters are obtained by minimizing the mismatch from ground truth data (with discrete index $i$) given by a 1 step loss function,
\begin{equation}
L_1=\mathbb{E}_i\left[\left\|F_{\theta}(q_i)-S(q_i)\right\|\right]
\end{equation}
The popular multi-rollout loss function, $L_M$ used to train several state-of-the-art models is defined as

\begin{equation}\label{Pfwd}
L_M=\mathbb{E}_i \left[\sum_{t=1}^{t=n}\left\|\lambda(t)(F_{\theta}^t(q_i)-S^t(q_i))\right\|\right]
\end{equation}
where $n$ is the number of rollouts that the training sees and $\lambda(t)$ is a hyper-parameter that gives lower weights to mismatch in trajectories that are farther in time. Furthermore, the 'Pushforward Trick' introduced in \cite{brandstetter2022message} can be used to reduce the computational cost and induce stability by breaking the computational graph between intermediate rollouts. However, these techniques are themselves insufficient to capture the invariant metric of the underlying dynamical system for prediction of chaotic systems (\cite{schiff2024dyslim}). The MP method introduces learnable intermediate discontinuities in the long trajectory and adds a penalty term to the rollout loss defined in Equation \ref{Pfwd} penalizing the magnitude of the discontinuities. Therefore, the problem of extreme gradients can be overcome and the trajectory learned is stable without explicitly specifying invariant properties in the loss function as in \cite{schiff2024dyslim} which are either unknown or computationally expensive to calculate.
\par
\begin{figure}[ht!]
    \centering
    \vspace{-0.65cm}
    \includegraphics[width=\linewidth]{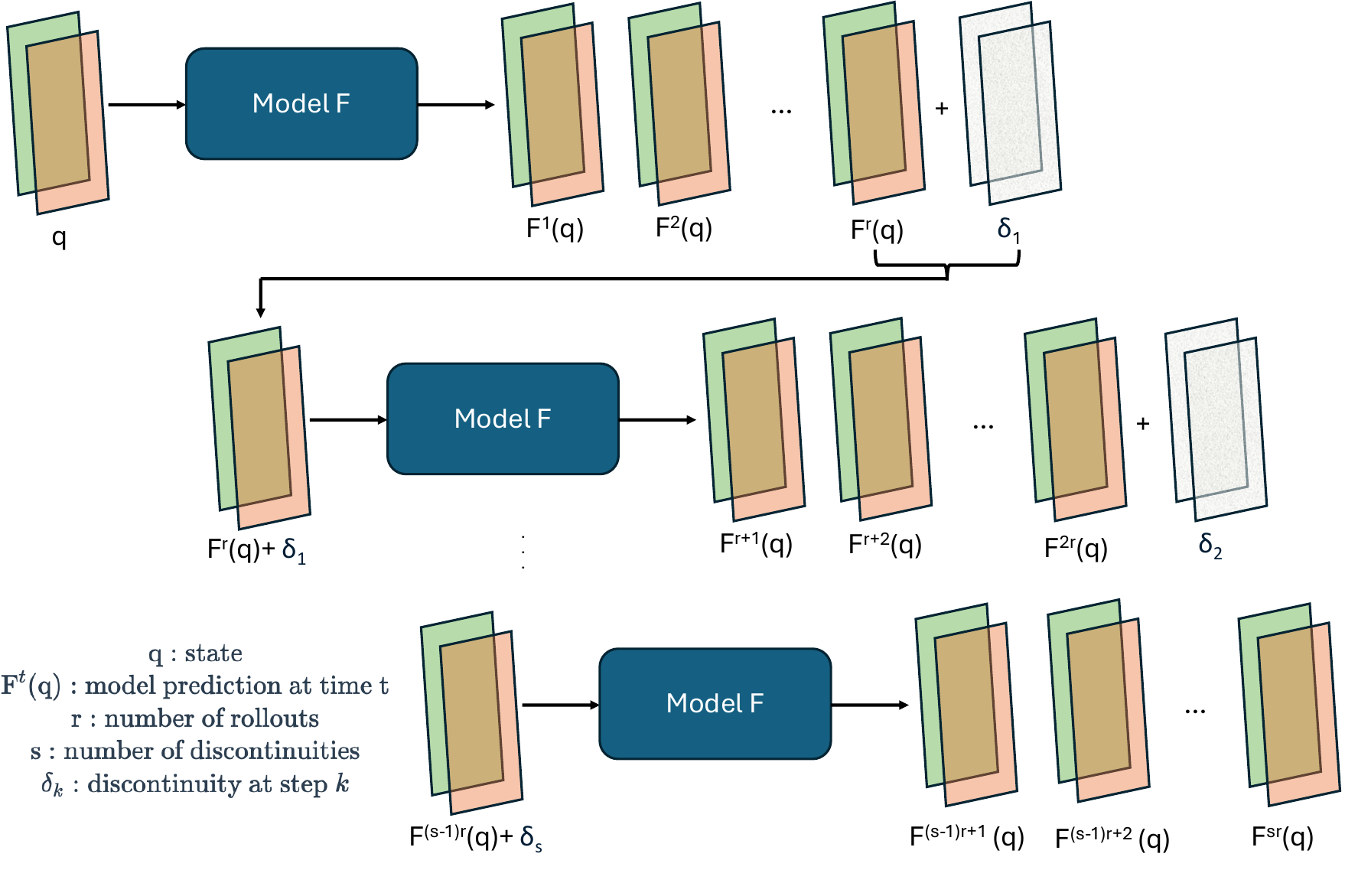}
    \vspace{-0.65cm}
    \caption{A schematic for MP optimization. The model $F$ can be any autoregressive machine learning model. The intermediate discontinuity $\delta$ is introduced after every $r$ rollouts.}
    \vspace{-0.5cm}
    \label{fig:MP}
\end{figure}
As shown in Figure \ref{fig:MP}, we augment the ground truth loss($L_{GT}$) with a penalty loss($L_P$) as,
\begin{equation}
    L_{MP}=\underbrace{\mathbb{E}_i \left[\sum_{t=1}^{t=sr}\left\|\lambda(t)(F_{\theta}^t(q_i)-S^t(q_i))\right\|\right]}_{L_{GT}} + \mu\underbrace{\sum_{k=1}^s \left\| \delta_k\right\|}_{L_P}
\end{equation}
where $r$ is the number of rollouts before introducing a discontinuity, $s$ is the number of splits (discontinuities) and $\delta$s are the introduced discontinuities (learnable parameters). This is a modification from the previous implementations of MP \citep{chung2022optimization,chakraborty2024divide} where the intermediate states were learnable. We found the proposed approach to be more scalable for larger systems and stable during training. However, after every $r$ rollouts we detached the computational graph before introducing the discontinuities $\delta$, so that gradients are not propagated through the entire trajectory. The penalty strength $\mu$ is a hyperparameter that is gradually increased to achieve continuity in time. It is typically started with a very low value($10^{-5}$ in our experiments) and then gradually increased. We also start with a single rollout (r=1) and a single discontinuity (s=1) in the trajectory which are gradually increased to learn longer trajectoies. Further details on tuning the hyperparameters of MP method are provided in \cite{chakraborty2024divide}.  The loss is backpropagated through the computational graph to compute the gradients with respect to the set of learnable parameters $\theta$s and $\delta$s using automatic differentiation. The intermediate discontinuities are introduced only in training and not used in inference. Techniques like the Pushforward Trick~\citep{brandstetter2022message} and weighting trajectories that are closer in time can also be used with the MP optimization. Any other improvement like the Maximum Mean Discrepancy (MMD) loss introduced in \cite{schiff2024dyslim} can be easily extended to MP method for further improvement. However, we note that the these additional invariant statistics based losses add significant computational overhead (requiring long-term integration for each gradient computation). The MP approach seeks to improve on standard autoregressive model training without this overhead.
\vspace{-0.4cm}
\section{Results}
\vspace{-0.25cm}
\subsection{Kolmogorov Flow}
\begin{figure}[ht!]
    \vspace{-0.55cm}
    \centering
    \includegraphics[width=0.95\linewidth]{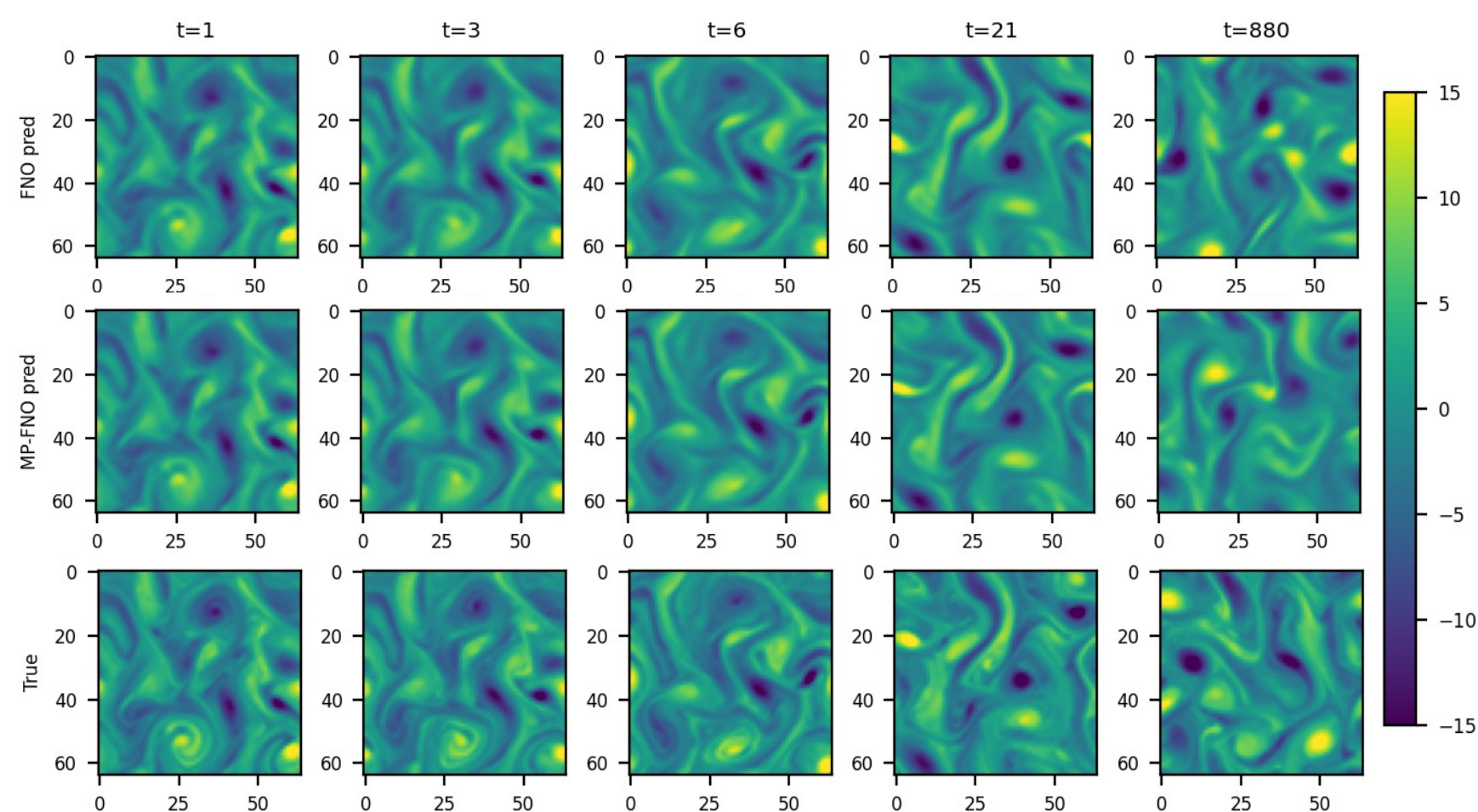}
    \vspace{-0.25cm}
    \caption{Vorticity of 2D Kolmogorov flow from predicted velocity fields. 't' here is the rollout step of the model.}
    \label{fig:KF}
    \vspace{-0.25cm}
\end{figure}
\begin{figure}[h]
    \centering
    \includegraphics[width=0.98\linewidth]{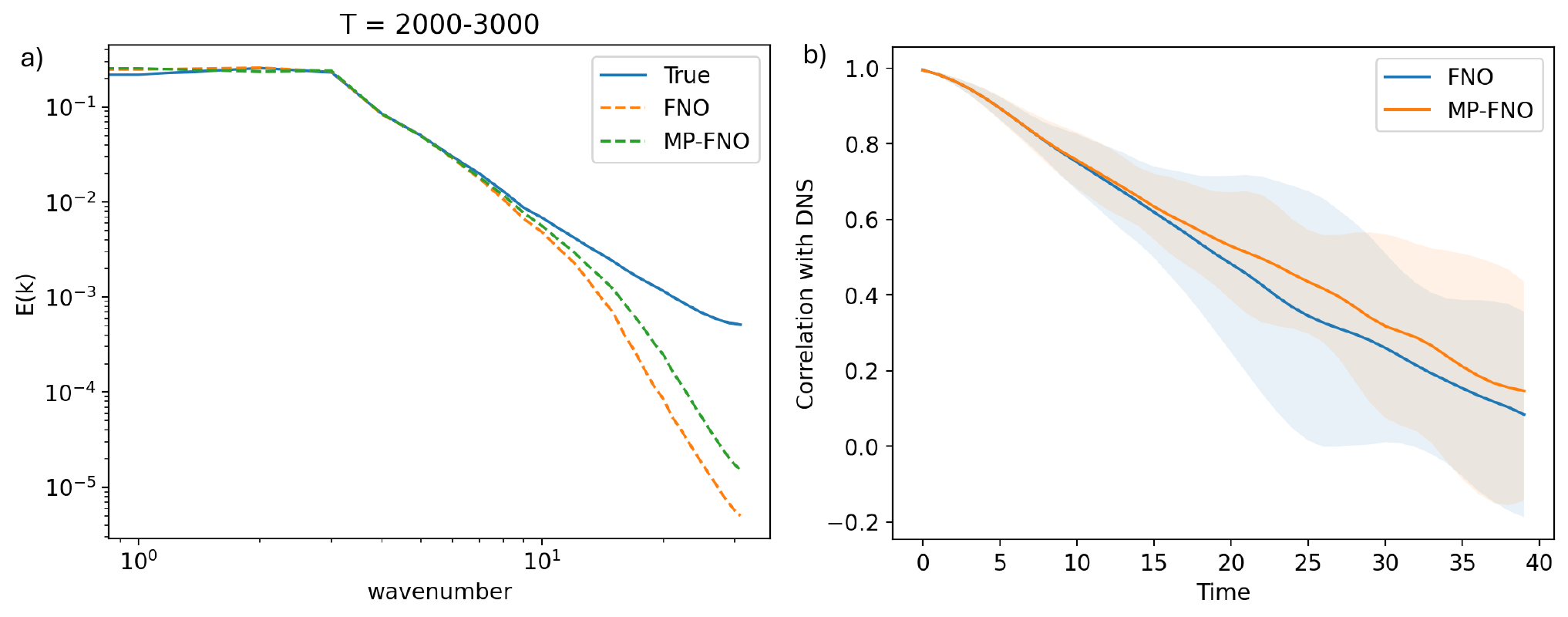}
    \caption{A comparison between FNO and MP-FNO for (a) angle-averaged total kinetic energy spectrum and (b) correlation with DNS. In (a) we check the performance for an invariant statistic, and for (b) we assess how the MP FNO technique improves accuracy with forecast duration compared to vanilla FNO. }
    \label{fig:kf_stats}
\end{figure}

\vspace{-0.2cm}
This section evaluates the performance of our proposed framework on two-dimensional homogeneous isotropic turbulence driven by Kolmogorov forcing, governed by the incompressible Navier-Stokes equations. These experiments aim to assess MP optimization's capabilities for improving performance of the Fourier Neural Operator. Forced two-dimensional turbulence, a classic example of chaotic dynamics, has become a standard benchmark for ML methods used in dynamical system prediction (\cite{stachenfeld2021learned,brandstetter2022message,schiff2024dyslim}). 
The Reynolds number $Re=10^5$ chosen for this study. The initial condition is a randomly generated divergence-free velocity field~\cite{Kochkov2021-ML-CFD}. For more details on dataset construction, refer to the work by \cite{shankar2023differentiable}. The trajectories are temporally sub-sampled empirically after flow reaches the chaotic regime to guarantee sufficient separation between snapshots. It can be observed in Figure \ref{fig:KF} that the the predictions match the ground truth closely in the starting timesteps and then diverge as a property of chaotic system. However, both FNO and MP-FNO shows no sign of instability even after a high number of autoregressive rollout.  
The MP optimization clearly improves upon the vanilla FNO for an invariant metric - the energy spectrum, and the correlation with DNS as shown in Figure \ref{fig:kf_stats}. The latter also demonstrates how the MP optimization improves accuracy with greater integration duration.
\vspace{-0.5cm}
\subsection{Ocean Reanalysis Data}
\vspace{-0.3cm}
\begin{figure}[!ht]
    \centering
    \includegraphics[width=0.9\linewidth]{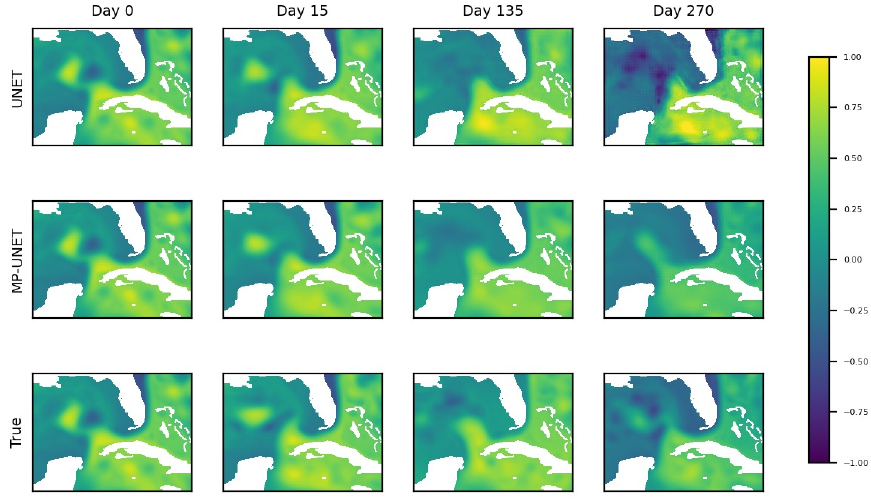}
    \vspace{-0.4cm}
    \caption{Prediction performance of UNET and MP-UNET for the GoM LCE shedding event: Eddy Sverdrup}
    \label{fig:sverdrup}
\end{figure}
In this experiment we implement the MP algorithm to predict the sea-surface height (SSH), longitudinal (SSU), and meridional (SSV) velocities of the northwest Atlantic Ocean’s western boundary extending from 92$^{\circ}$W into the Atlantic 75$^{\circ}$W in the Gulf of Mexico (GoM). For this, we have used the GLORYS version 4~\citep{garcia2021introduction} reanalysis dataset, which is an eddy-permitting dataset at $\frac{1}{12}^{\circ}$ ($8$ Km). The training data (available daily) is temporally sub-sampled by a factor of 3 to keep sufficient distinction between the snapshots. We implement the MP algorithm with a UNET \citep{ronneberger2015u} architecture and compare the predictions for a test (unseen) time period of a major GoM Loop Current Eddies (LCE) shedding event: Eddy Sverdrup (Jul 2019-Jan 2020).
\par
\begin{figure}[!h]
    \centering
        \includegraphics[width=0.6\linewidth]{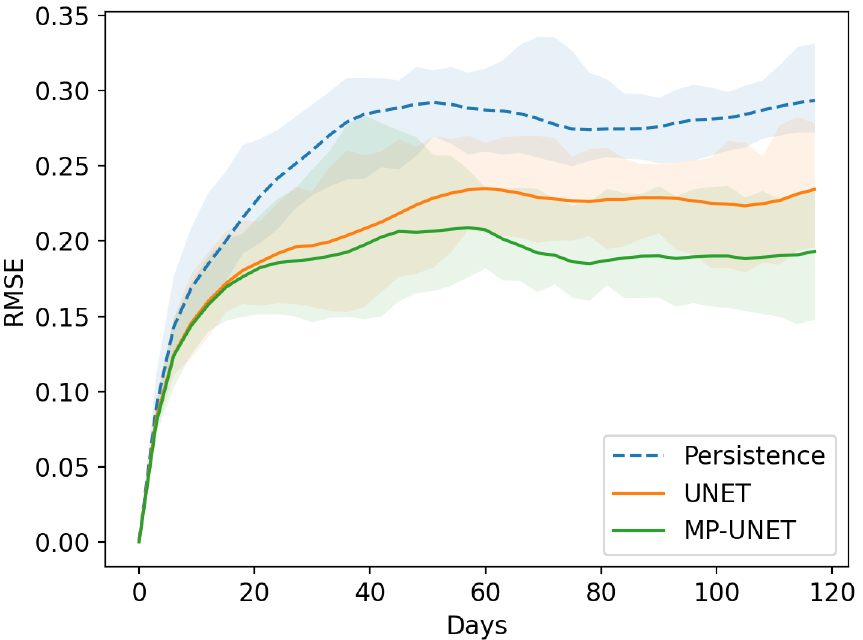}
    \caption{RMSE comparison between UNET, MP-UNET and Persistence. Persistence is an elementary model used to compare the performance of other models. It assumes that the weather is static and the initial condition itself is the forecast.
    }
    \label{fig:RMSE_sverdrup}
\end{figure}

Figure \ref{fig:sverdrup} shows that both UNET and MP-UNET captures the dynamics of the data accurately for the time-period of the eddy event. We also found out that the vanilla UNET shows signs of instability after longer periods of time whereas MP optimization makes it stable for over 270 days while testing. However, to delve deeper into the results we compare root-mean-square error (RMSE) from ground truth for the model predictions. MP-UNET performs the best in long term as evident from Figure \ref{fig:RMSE_sverdrup}.

This also demonstrates the potential for the MP technique to improve neural forecasting applications in real-world use cases, for example in the earth sciences \citep{kashinath2021physics}.
\vspace{-0.2cm}
\section{Conclusion}
\vspace{-0.2cm}
This paper focuses on the challenges posed by the long-term prediction of chaotic systems. Our proposed method provides a modified extension of the multi-step penalty(MP) optimization framework to a broader class of deep learning models such as Fourier Neural Operators and UNETs. We demonstrate its advantage by forecasting challenging chaotic systems such as high Reynolds number 2D turbulence and the Gulf Stream ocean reanalysis dataset. The MP optimization based architectures show more stability in the long term and is more accurate in short term compared to their vanilla counterparts without any significant overhead in computational cost. This work contributes to the field of data-driven modeling of chaotic systems and opens new avenues to explore and gain insight into complex natural phenomena.

\section*{Acknowledgements}
This material is based upon work supported by the U.S. Department of Energy (DOE), Office of Science, Office of Advanced Scientific Computing Research, under Contract DE-AC02-06CH11357. This research was funded in part and used resources of the Argonne Leadership Computing Facility, which is a DOE Office of Science User Facility supported under Contract DE-AC02-06CH11357. RM and DC acknowledge support from U.S. Department of Energy, Office of Science, Office of Advanced Scientific Computing Research grant DOE-FOA-2493: ``Data-intensive scientific machine learning'', and computational support from Penn-State Institute for Computational and Data Sciences.

\bibliography{references}

\end{document}